\documentclass[10pt,twocolumn,letterpaper]{article}

\usepackage{cvpr}
\usepackage{times}
\usepackage{epsfig}
\usepackage{graphicx}
\usepackage{amsmath}
\usepackage{amssymb}
\usepackage{amsfonts}
\usepackage{subcaption}
\usepackage{tabularx}
\usepackage{bbm}

\usepackage{mathtools}
\usepackage{multirow}
\makeatletter

\graphicspath{{images/}}

\def\BState{\State\hskip-\ALG@thistlm}

\newcommand{\comment}[1]{}
\newcommand{\parag}[1]{\vspace{-3mm}\paragraph{#1}}

\newif\ifdraft
\drafttrue

\usepackage{booktabs}
\usepackage[pagebackref=true,breaklinks=true,letterpaper=true,colorlinks,bookmarks=false]{hyperref}

\newcommand{\phantomone}{\phantom{a}}
\newcommand{\phantomtwo}{\phantom{aa}}

\newcommand{\ourS}[0]{\emph{Ours-SRU}}
\newcommand{\ourD}[0]{\emph{Ours-DRU}}
\newcommand{\refN}[0]{\emph{RefineNet}}
\newcommand{\recS}[0]{\emph{Rec-Simple}}
\newcommand{\recL}[0]{\emph{Rec-Last}}
\newcommand{\recM}[0]{\emph{Rec-Middle}}
\newcommand{\Unet}[0]{\emph{U-Net}}

\newcommand{\TourS}[0]{{Ours-SRU}}
\newcommand{\TourD}[0]{{Ours-DRU}}
\newcommand{\TrefN}[0]{{RefineNet}}
\newcommand{\TrecS}[0]{{Rec-Simple}}
\newcommand{\TrecL}[0]{{Rec-Last}}
\newcommand{\TrecM}[0]{{Rec-Middle}}
\newcommand{\TUnet}[0]{{U-Net}}

 \cvprfinalcopy 
\def\cvprPaperID{***} 

\ifcvprfinal\fi

\begin{document}
\title{Beyond One Glance: Gated Recurrent Architecture for Hand Segmentation}

\author{Wei Wang\thanks{The first two authors contribute equally.} \and Kaicheng Yu\footnotemark[1] \and Joachim Hugonot \and Pascal Fua \and Mathieu Salzmann\thanks{This work was supported in part by the Swiss Innovation Agency Innosuisse and by the Swiss National Science Foundation.
	} \\
	Computer Vision Laboratory, \'Ecole polytechnique f\'ed\'erale de Lausanne~(EPFL) \\
{\tt\small \{wei.wang, kaicheng.yu, joachim.hugonot, pascal.fua, mathieu.salzmann\}@epfl.ch}
}

\maketitle


\begin{abstract}

	As mixed reality is gaining increased momentum, the development of effective and efficient solutions to egocentric hand segmentation is becoming critical. Traditional segmentation techniques typically follow a one-shot approach, where the image is passed forward only once through a model that produces a segmentation mask. This strategy, however, does not reflect the perception of humans, who continuously refine their representation of the world. 
	
	In this paper, we therefore introduce a novel gated recurrent architecture. It goes beyond both iteratively passing the predicted segmentation mask through the network and adding a standard recurrent unit to it. Instead, it incorporates multiple encoder-decoder layer of the segmentation network, so as to keep track of its internal state in the refinement process.
	As evidenced by our results on standard hand segmentation benchmarks and on our own dataset, our approach outperforms these other, simpler recurrent segmentation techniques, as well as the state-of-the-art hand segmentation one. Furthermore, we demonstrate the generality of our approach by applying it to road segmentation, where it also outperforms other baseline methods.

\end{abstract}

\section{Introduction}

With the growing popularity of wearable cameras and augmented reality (AR), egocentric hand analysis has received increasing attention. Understanding hand motion would make it possible to interact with both virtual objects in an augmented reality context and real ones in an augmented virtuality setup. Of particular interest is hand-segmentation that enables, among other things,  the insertion of the user's real hands in a virtual environment. This is challenging because highly accurate segmentations are required for a convincing illusion, even though the hand motion can be very complex. 


\begin{figure}[t]
	\begin{center}
		\includegraphics[width=\linewidth]{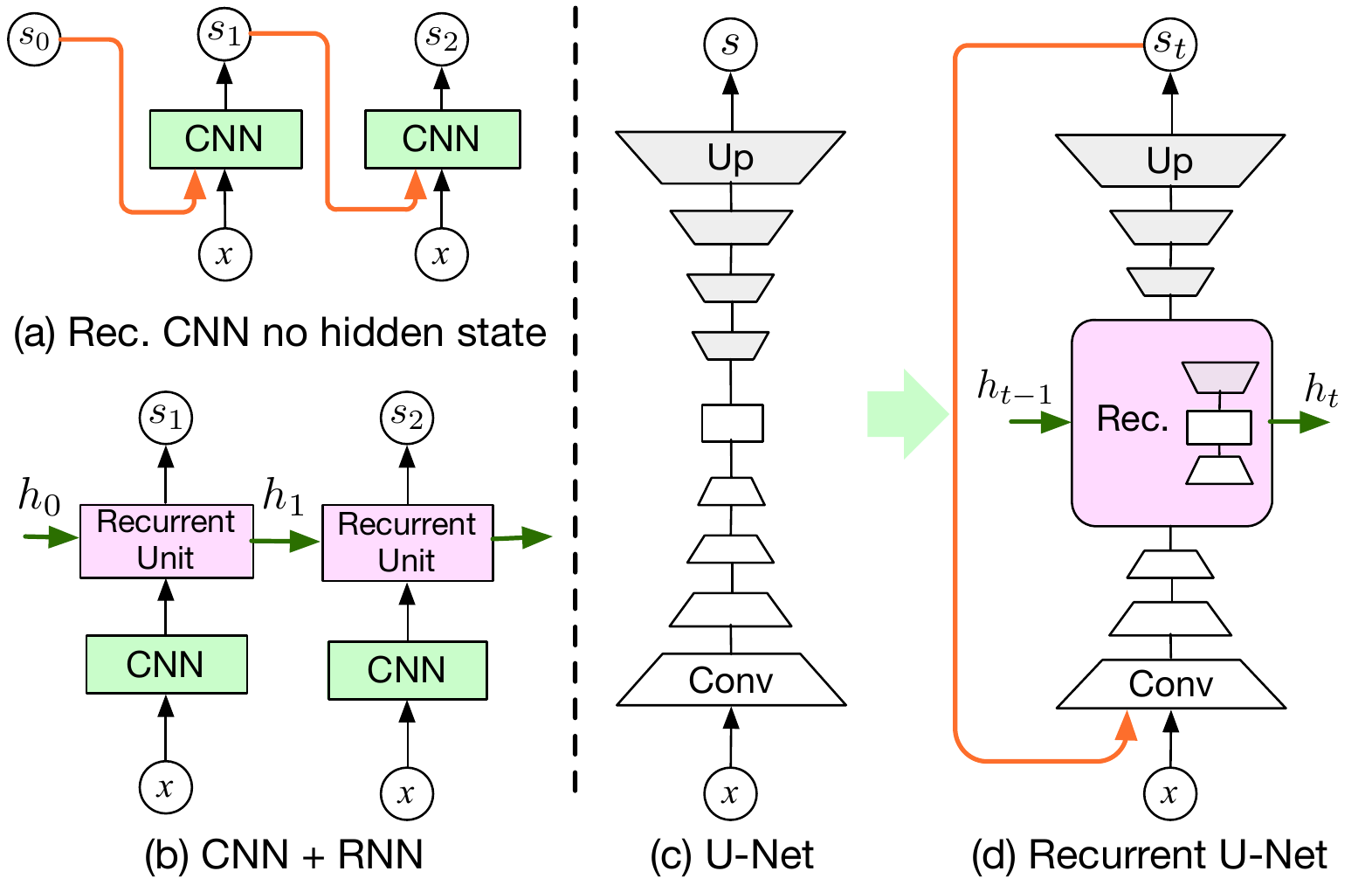}
	\end{center}
	\vspace{-3mm}
	\caption{{\bf Recurrent Segmentation.} {\bf (a)} The simple strategy of~\cite{Mosinska18,Pinheiro14} consists of concatenating the previous segmentation mask $s_{t-1}$ to the image $x$, and recurrently feeding this to the network. {\bf (b)} For sequence segmentation, to account for the network's internal state, one can instead combine the CNN with a standard recurrent unit as in~\cite{Valipour17}. Here, we build upon the U-Net architecture of~\cite{Ronneberger15} {\bf  (c)}, and propose to build a recurrent unit over several of its layers, as shown in {\bf (d)}. This allows us to propagate higher-level information through the recurrence, and, in conjunction with a recurrence on the segmentation mask, outperforms the two simpler recurrent architectures  {\bf (a)} and {\bf (b)}.
	}
	\label{fig:rcnn}
\end{figure}

Most recent image segmentation work~\cite{Lin17c,Ronneberger15,Long15a}, whether for hands or for other objects, has focused on designing \emph{one-shot} methods, that is, algorithms that take the image as input, process it, and return an output without any feedback loop. This is in contrast with what we know about human vision. According to neuroscience research~\cite{Purves01}, when we observe a scene, our eyes undergo saccadic movements, and we accumulate knowledge about the scene and continuously refine our perception. Our motivation is therefore to mimic human perception by introducing a new recurrent architecture and thereby increase segmentation performance. 

Earlier refinement strategies that operate along those lines~\cite{Pinheiro14,Mosinska18} perform an initial segmentation and then iteratively take as input the segmentation mask computed at the previous iteration and the image to produce a new refined mask. This is illustrated by Fig.~\ref{fig:rcnn}(a). A drawback of this approach is that it does not keep track of the network's internal state during the refinement process.

In this paper, we introduce a novel more sophisticated recurrent segmentation architecture. It iteratively refines both the segmentation mask and the network's internal state. To this end, we build upon the encoder-decoder U-Net architecture of~\cite{Ronneberger15}, depicted by Fig.~\ref{fig:rcnn}(c). As shown in Fig.~\ref{fig:rcnn}(d), we retain its overall structure but build a recurrent unit over some of its inner layers for internal state update. By contrast with the simple CNN+RNN architecture of Fig.~\ref{fig:rcnn}(b), often used for video or volumetric segmentation~\cite{Valipour17,Poudel16,Ballas16}, this allows the network to keep track of and to iteratively update more than just a single-layer internal state. In fact, this architecture gives us the flexibility to choose the portion of the internal state that we exploit for recursion purposes and to explore variations of our overall scheme. 

We will demonstrate that our method outperforms earlier and simpler approaches to recursive segmentation~\cite{Mosinska18} as well as the state-of-the-art RefinetNet-based method of~\cite{Urooj18} on several modern egocentric hand segmentation datasets. They include the GTEA dataset~\cite{Fathi11}, the Ego-hand dataset~\cite{Bambach15},  the EgoYouTube-Hands (EYTH) dataset~\cite{Urooj18}, and the HandOverFace (HOF) dataset~\cite{Urooj18}. As these publicly available datasets are relatively small, with at most 4.8K annotated images, we demonstrate the scalability of our approach, along with its applicability in a keyboard typing scenario, by introducing a large dataset containing 12.5K annotated images. This dataset and our code will be made publicly available upon acceptance of the paper.

Our contributions are therefore a novel approach to recurrent image segmentation that outperforms state-of-the-art hand-segmentation methods, along with a new large-scale hand-segmentation dataset. Furthermore, our approach is generic and can be applied to other segmentation tasks. To demonstrate this, we tested it on the same road dataset as the one used in~\cite{Mosinska18} and show that it also outperforms other recurrent methods in a context different from the one we designed our model for. 


\section{Related Work}

\subsection{Hand Segmentation}

Hand segmentation is an old Computer Vision problem that has received much attention over the years. Traditional methods can be grouped into those that rely on local appearance~\cite{Jones02,Kakumanu07,Argyros04,Kolsch05}, those that leverage a hand template~\cite{Sudderth04,Stenger01a,Oikonomidis10}, and those that exploit motion~\cite{Sheikh09,Hayman03,Fathi11}. 
As in many other areas, however, most state-of-the-art approaches now rely on deep networks~\cite{Li13d,Bambach15,Betancourt17,Urooj18}. In particular, the recent work of~\cite{Urooj18} that relies on a RefineNet~\cite{Lin17c} constitutes the current state of the art.

The RefineNet~\cite{Lin17c}, as most other semantic segmentation networks, such as U-Nets~\cite{Ronneberger15}, and FCNs~\cite{Long15a}, performs {\it one shot} segmentation. That is, the source image is passed only once through the network, which directly outputs the segmentation map. Here, however, motivated by the saccadic movements in human perception that continuously refine our representation of the world, we argue and empirically demonstrate that segmentation should be a recursive process. Note that building a recurrent approach over the RefineNet would have been impractical because of its large size. We therefore built it upon the U-Net, which has proven effective.

\subsection{Recurrent Networks for Segmentation}

While the main trend in semantic segmentation remains the one-shot approach, recurrent networks have nevertheless been studied in this context. For example, in~\cite{Mosinska18}, the segmentation mask produced by a modified U-Net was passed back as input to it along with the original image, which resulted in a progressive refinement of the segmentation mask. This approach is illustrated in Fig.~\ref{fig:rcnn}(a). A similar approach was also followed in the earlier work of~\cite{Pinheiro14}, where the resolution of the input image patch varied across the iterations of the refinement process. Instead of including the entire network in the recursive procedure, another approach consists of adding a standard recurrent unit at the output of the segmentation network, as shown in Fig.~\ref{fig:rcnn}(b). This was for instance employed in~\cite{Romera16} to iteratively produce the segmentation mask of different object instances in the scene. In principle, however, such a standard convolutional recurrent unit~\cite{Ballas16,Poudel16,Valipour17} could also be directly used for iterative segmentation of a single object.

In this paper, we introduce a more sophisticated recurrent segmentation method grounded on the premise that one should not only iteratively exploit the segmentation mask as input to the network, but also keep track of its internal state. To this end, we develop the architecture depicted by Fig.~\ref{fig:rcnn}(d) in which part of the segmentation network is contained in the recurrent unit. Our experiments demonstrate that this approach outperforms the state-of-the-art hand segmentation techniques, as well as the two types of recurrent segmentation methods discussed above.


\section{Method}

We now introduce our novel recurrent semantic segmentation architecture. To this end, we first discuss the overall structure of our framework, and then provide the details of the recurrent unit it relies on. Finally, we briefly discuss the training strategy for our approach.

\subsection{Recurrent U-Net}
\label{sec:r-unet}

\begin{figure*}[t]
	\includegraphics[width=\textwidth]{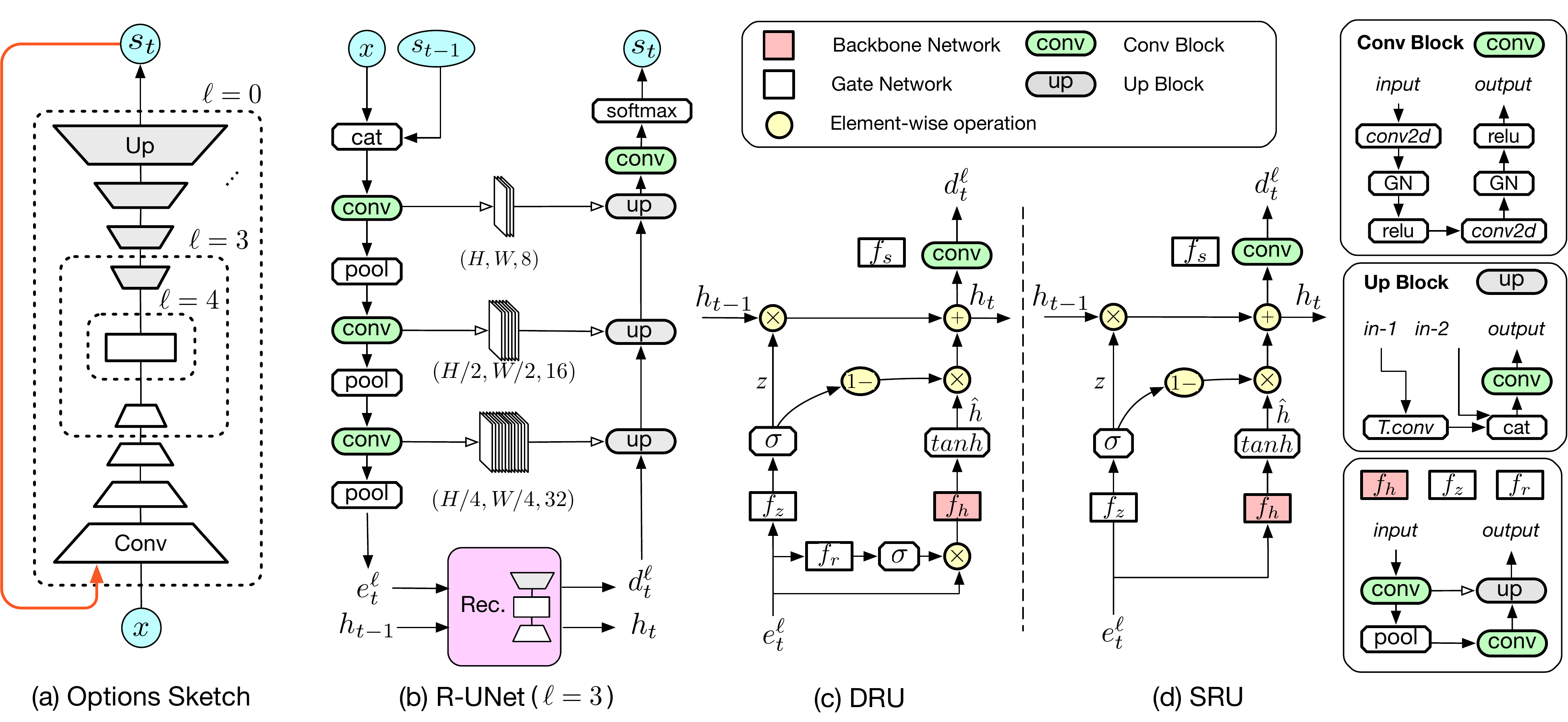}

	\caption{\textbf{Recurrent UNet (R-UNet).} {\bf (a)} As illustrated in Fig.~\protect\ref{fig:rcnn}(d), our model incorporate several encoding and decoding layers in a recurrent unit. The choice of which layers to englobe is defined by the parameter $\ell$. {\bf (b)} For $\ell = 3$, the recurrence occurs after the third pooling layer in the U-Net encoder. The output of the recurrent unit is then passed through three decoding up-convolution blocks. We design two different recurrent units, the Dual-gated Recurrent Unit (DRU)~\textbf{(c)} and the Single-gated Recurrent Unit (SRU)~\textbf{(d)}. They differ by the fact that the first one has an additional reset gate acting on its input. See the main text for more detail. }
	\label{fig:model_arch}
\end{figure*}

We rely on the U-Net architecture of~\cite{Ronneberger15} as backbone to our approach. As shown in Fig.~\ref{fig:model_arch}(a), the U-Net has an encoder-decoder structure, with skip connections between the corresponding encoding and decoding layers that allow the network to retain low-level features for the final prediction.
In practice, the datasets we work with are not particularly large. We therefore rely on a U-Net design where the first convolutional unit has 8 feature channels, and, following the original U-Net strategy, the channel number doubles after every pooling layer in the encoder. The decoder relies on transposed convolutions to increase the model's representation power compared to bilinear interpolation. We use group-normalization~\cite{Wu18a} in all convolutional layers since the batch size is usually very small.

Our contribution is to integrate recursions on the predicted segmentation mask $s$ and on multiple internal states of the network. The former can be achieved by simply concatenating, at each recurrent iteration $t$, the previous segmentation mask $s_{t-1}$ to the input image, and passing the resulting tensor through the network. For the latter, we propose to replace a subset of the encoding and decoding layers of the U-Net with a recurrent unit. Below, we first formalize this unit, and then discuss two variants of its internal mechanism. 

To formalize our recurrent unit, let us consider the process at iteration $t$ of the recurrence. At this point, the network takes as input an image $x$ concatenated with the previously-predicted segmentation mask $s_{t-1}$. Let us then denote by $e^{\ell}$ the activations of the $\ell^{th}$ encoding layer, and by $d_{\ell}$ those of the corresponding decoding layer. Our recurrent unit takes as input $e^{\ell}$, together with its own previous hidden tensor $h_{t-1}$, and outputs the corresponding activations $d_{\ell}$, along with the new hidden tensor $h_t$. Note that, to mimic the computation of the U-Net, we use multiple encoding and decoding layers within the recurrent unit.

In practice, one can choose the specific level $\ell$ at which the recurrent unit kicks in. In Fig.~\ref{fig:model_arch}~(b), we illustrate the whole process for $\ell=3$. 
When $\ell=0$, the entire U-Net is included in the recurrent unit, which then takes the concatenation of the segmentation mask and the image as input. Note that, for $\ell = 4$, the recurrent unit still contains several layers because the central portion of the U-Net in Fig.~\ref{fig:model_arch}(a) corresponds to a convolutional {\it block}.
In our experiments, we evaluate two different structures for the recurrent units, which we discuss below.

\subsection{Dual-gated Recurrent Unit}
\label{sec:dru}

As a first recurrent architecture, we draw inspiration from the Gated Recurrent Unit (GRU)~\cite{Cho14c}. As noted above, however, our recurrent unit replaces multiple encoding and decoding layers of the segmentation network. We therefore modify the equations accordingly, but preserve the underlying motivation of GRUs. Our architecture is shown in Fig.~\ref{fig:model_arch}(c).

Specifically, at iteration $t$, given the activations $e_{\ell}^t$ and the previous hidden state $h_{t-1}$, we aim to produce a candidate update $\hat{h}$ for the hidden state and combine it with the previous one according to how reliable the different elements of this previous hidden state tensor are. In our context, to determine this reliability, we make use of an update gate defined by a tensor 
\begin{equation}
z = \sigma( f_{z}(e^{\ell}_{t}) )\;,
\label{eq:p}
\end{equation}
where $f_{z}(\cdot)$ denotes an encoder-decoder network with the same architecture as the portion of the U-Net that we replace with our recurrent unit.

Similarly, we obtain the candidate update as
\begin{equation}
\hat{h} = tanh(f_{h}(r\odot e^{\ell}_{t}))\;,
\label{eq:q}
\end{equation}
where $f_{h}(\cdot)$ is a network with the same architecture as $f_{z}(\cdot)$, but a separate set of parameters, $\odot$ denotes the element-wise product, and $r$ is a reset tensor allowing us to mask parts of the input used to compute $\hat{h}$. This reset tensor is computed as
\begin{equation}
r = \sigma( f_{r}(e^{\ell}_{t}) )\;,
\end{equation}
where $f_r(\cdot)$ is again a network with the same encoder-decoder architecture as before.

Given these different tensors, the new hidden state is computed as
\begin{equation}
h_t = z \odot h_{t-1} + (1-z) \odot \hat{h}\;.
\label{eq:ht}
\end{equation}
Finally, we predict the output of the recurrent unit, which corresponds to the activations of the $\ell^{th}$ decoding layer as
\begin{equation}
d^{\ell}_{t} = f_{s}(h_{t})\;,
\label{eq:st}
\end{equation}
where, as shown in Fig.~\ref{fig:model_arch}(c), $f_s(\cdot)$ is a simple convolutional block.

Since it relies on two gates, $r$ and $z$, we dub this recurrent architecture Dual-gated Recurrent Unit (DRU). 
One main difference with GRUs is the fact that we use multi-layer encoder-decoder networks in the inner operations instead of simple linear layers. Furthermore, note that, in contrast to GRUs, we do not directly make use of the hidden state $h_{t-1}$ in these inner computations. This allows us not to have to increase the number of channels in the encoding and decoding layers compared to the original U-Net. Nevertheless, the hidden state is indirectly employed, since, via the recursion, $e^{\ell}_t$ depends on $d^{\ell}_{t-1}$, which is computed from $h_{t-1}$.

\subsection{Single-Gated Recurrent Unit}
\label{sec:sru}

As evidenced by our experiments, the DRU described above is effective at iteratively refining a segmentation. However, it suffers from the drawback that it incorporates three encoder-decoder networks, which may become memory-intensive depending on the choice of $\ell$. To decrease this cost, we therefore introduce a simplified recurrent unit, which relies on a single gate, thus dubbed Single-gated Recurrent Unit (SRU).

Specifically, as illustrated in Fig.~\ref{fig:model_arch}(d), our SRU has a structure similar to that of the DRU, but without the reset tensor $r$. As such, the equations remain mostly the same as above, with the exception of the candidate hidden state, which we now express as

\begin{equation}
\hat{h} = tanh(f_{h}(e^{\ell}_{t}))\;.
\label{eq:q2}
\end{equation}
This simple modification allows us to remove one of the encoder-decoder networks from the recurrent unit, which, as shown by our results, comes at very little loss in segmentation accuracy.

\subsection{Training}
\label{sec:loss}

To train our recurrent U-Net, we use the cross-entropy loss. More specifically, we introduce supervision at each iteration of the recurrence. To this end, we write our overall loss as
\begin{equation}
L = \sum_{t=1}^N w_t L_t\;,
\end{equation}
where $N$ represents the number of recursions, set to 3 in this paper, and $L_t$ denotes the cross-entropy loss at iteration $t$, which is weighted by $w_t$. We define this weight as
\begin{equation}
w_t = \alpha^{N-t} \;,
\label{eq:rel_weight}
\end{equation}
which, by setting $\alpha \leq1$, increases monotonically with the iterations.  In our experiments, we either set $\alpha = 1$, so that all iterations have equal importance, or $\alpha = 0.4$, thus encoding the intuition that we seek to put more emphasis on the final prediction. A detailed study of the influence of $\alpha$ is provided in Section~\ref{sec:abl_para}. 


\section{Dataset}

In our experiments, we demonstrate the benefits of our approach on standard benchmark datasets, such as GTEA~\cite{Fathi11}, EYTH~\cite{Urooj18}, EgoHand~\cite{Bambach15}, and HOF~\cite{Urooj18}. These datasets, however, are relatively small, with at most 4,800 images in total, as can be seen in Table~\ref{tab:dsComp}. 

\begin{figure}[t]
  \includegraphics[width=\linewidth]{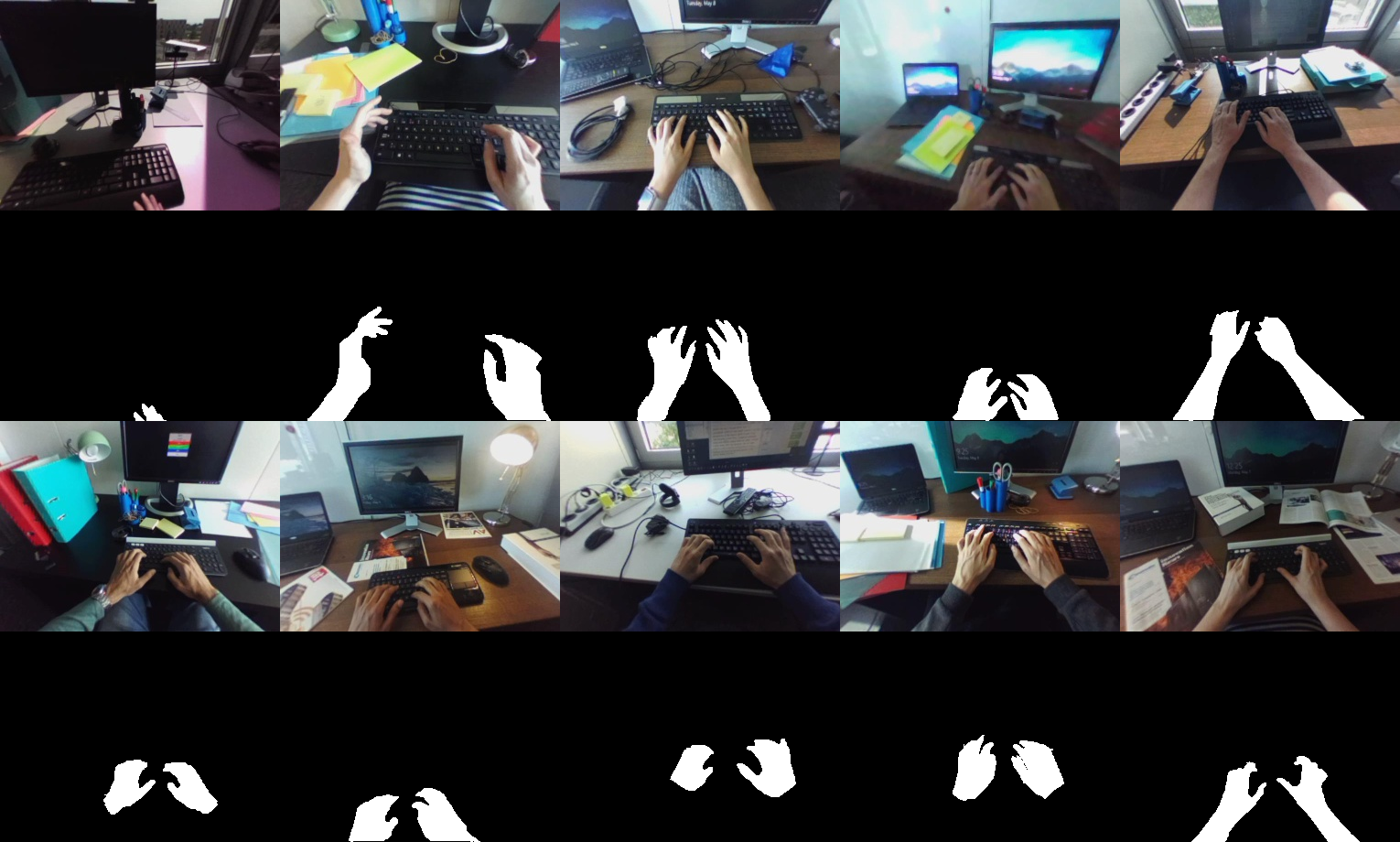}
  \caption{\small \textbf{Keyboard Hand (KBH) dataset.} Sample images featuring diverse environmental and lighting conditions, along with associated ground-truth segmentations.}
  \label{fig:annotEx}
\end{figure}

To evaluate our approach on a larger-scale one, we therefore acquired our own. Because this work was initially motivated by an augmented virtuality project whose goal is to allow someone to type on a keyboard while wearing a head-mounted display, we asked 50 people to type on 9 different keyboards while wearing an HTC Vive~\cite{HTCvive}. To make this easier, we created a mixed-reality application to allow the users to see both the camera view and a virtual browser showing the text being typed. 
To ensure diversity, we varied the keyboard types, lighting conditions, desk colors, and objects lying on them, as can be seen in Fig.~\ref{fig:annotEx}.  We provide additional details in Table~\ref{tab:kbh}. 

\begin{table}[!t]
	\centering
	\resizebox{0.9\linewidth}{!}{%
	\setlength{\tabcolsep}{3pt}
		\begin{tabular}[width=\textwidth]{l c  @{}rl c @{}rrrr  c}
			\toprule
			 & \phantomone & \multicolumn{2}{c}{Resolution} & \phantomone& \multicolumn{4}{c}{\# Images} \\
			\cmidrule(r){3-4}  	\cmidrule(r){5-9} 
			Dataset && Width & Height &&   Train & Val. & Test & Total \\

			\midrule 
		KBH (Ours)&& $230{\times}$&$306$ && 2300 & 2300 & 7936 & 12536 \\
			\midrule 
		EYTH~\cite{Urooj18} && $216{\times}$&$384$ && 774 & 258 & 258 &1290 \\
		HOF~\cite{Urooj18} && $216{\times}$&$384$ && 198&40&62 & 300 \\
		EgoHand~\cite{Bambach15} && $720{\times}$&$1280$ && 3600&400&800 & 4800 \\
		GTEA\cite{Fathi11} && $405{\times}$&$720$ && 367 & 92 & 204 & 663\\
		\bottomrule
		\end{tabular}
	}
	\caption{\small Hand-segmentation benchmark datasets.}
		\label{tab:dsComp}
\end{table}

\begin{table}[!t]
	\setlength{\tabcolsep}{2pt}
	\resizebox{\linewidth}{!}{
		\begin{subtable}{0.5\linewidth}
			\centering
			\label{tab:ds_params}
			\resizebox{!}{45pt}{%
				\begin{tabular}{ccc}
					\multicolumn{3}{c}{(a) Environment setup}\\
					\toprule
					Parameters & Amount & Details \\
					\midrule
					Desk & 3 & White, Brown, Black\\
					Desk position & 3 & - \\
					Keyboard & 9 & -\\
					Lighting & 8 & 3 sources on/off\\
					Objects on desk & 3 & 3 different objects\\
					\bottomrule
				\end{tabular}
			}
		\end{subtable}%
		\begin{subtable}{.3\linewidth}
			\resizebox{!}{45pt}{%
				\begin{tabular}{c}
					\phantomtwo \\
				\end{tabular}%
			}
		\end{subtable} 
		\begin{subtable}{0.25\linewidth}
			\centering
			\label{tab:ds_attrib}
			\resizebox{!}{45pt}{%
				\begin{tabular}{cc}
					\multicolumn{2}{c}{(b) Attributes} \\
					\toprule
					Attribute & \#IDs \\
					\midrule
					Bracelet & 10\\
					Watch & 14\\
					Brown-skin & 2\\
					Tatoo & 1\\
					Nail-polish & 1\\
					Ring(s) & 6\\
					\bottomrule
				\end{tabular}%
			}
		\end{subtable} 
	}
	\caption{\small Properties of our new KBH dataset.}
	\label{tab:kbh}
\end{table}

We then recorded 161 hand sequences with the device's camera.   We split them as 20-20-60\% for Training/Validation/Testing to setup a challenging scenario in which the training data is not overabundant and to test the scalability and generalizability of the trained models. We guaranteed that the same person never appears in more than one of these splits by using people's IDs during partitioning. In other words, our splits resulted in three groups of 30, 30, and 101 separate videos, respectively. 

We annotated about the same number of frames in each one of the videos, resulting in a total of 12,536 annotated frames. To this end, we developed a tool based on OpenCV's implementation of Grabcut~\cite{Rother04}. In practice, the annotator provides a few hard labels, in the form of short segments drawn with the mouse, and an initial segmentation is obtained using the graph-cut algorithm. The annotator can then iteratively refine this segmentation by incorporating new hard labels. Fig.~\ref{fig:annotEx} depicts the resulting ground-truth segmentations in a few frames. 

\section{Experiments}

In this section, we first validate the parameters of the two versions of our Recurrent U-Net. We then compare it against the state of the art on publicly available hand-segmentation benchmark datasets and on the larger one we introduced in the previous section. Finally, to demonstrate the generality of our approach, we evaluate it for road segmentation purposes. 

\subsection{Experimental Setup}

We now discuss the baselines, metrics, and experimental protocols we use to compare our approach to others. 

\parag{Baselines.}

We will refer to the version of our approach that relies on the dual gated unit of Section~\ref{sec:dru} and the single gated unit of  Section~\ref{sec:sru} as \ourS{} and \ourD{}, respectively. We compare them against the following baselines.

\begin{itemize}

\item  {\bf \refN{}}~\cite{Urooj18}. The current state-of-the-art hand segmentation method that relies on RefineNet. On the standard benchmarks, we directly report the numbers from~\cite{Urooj18}. For our new dataset, we adopt the same training procedure as in~\cite{Lin17c}, also employed in~\cite{Urooj18}, which uses the pre-trained ResNet-101 on the Pascal Human-parts dataset.

\item {\bf \Unet-B} and {\bf \Unet-G}~\cite{Ronneberger15}. We treat our U-Net backbone by itself as a baseline. \Unet-B{} uses batch-normalization and \Unet-G{}  group-normalization. For a fair comparison, they, \ourS{}, \ourD{}, and the recurrent baselines introduced below all use the same parameter settings.

\item {\bf \recL}. It has been proposed to add a recurrent unit after a convolutional segmentation network to process sequential data, such as video~\cite{Poudel16}. The corresponding \Unet-based architecture can be directly applied to segmentation by inputing the same image at all time steps, as shown in Fig.~\ref{fig:rcnn}(b). The output then evolves as the hidden state is updated. 

\item {\bf \recM }. Similarly, the recurrent unit can replace the bottleneck between the U-Net encoder and decoder, instead  of being added at the end of the network. This has been demonstrated to handle volumetric data~\cite{Valipour17}. Here we test it for segmentation purposes. The hidden state then is of the same size as the inner feature backbone, that is, 128 in our  experimental setup. 

\item {\bf \recS}~\cite{Mosinska18}. We perform a recursive refinement process, that is, we  concatenate the segmentation mask with the input image and feed it into the network. Note that the original method of~\cite{Mosinska18} relies on a VGG-19 pre-trained on ImageNet~\cite{Simonyan15}, which is far larger than our \Unet. To make the comparison fair, we therefore implement this baseline with the same U-Net backbone as in our approach.

\end{itemize}

\parag{Training Protocol.}

We use standard stochastic gradient descent with momentum 0.9 as the optimizer. Since different recurrent structures behave differently during training, we tune the learning rate for each model in the range 1e-3 to 1e-9. Furthermore, since the gradients obtained by summing over all pixels can become quite large, we clip the gradients of magnitudes greater than 10 for all recurrent approaches to prevent gradient explosion~\cite{Goodfellow16}. For all U-Net-based methods, the parameters are initialized randomly as in~\cite{He15}. For all recurrent models, we use the multi-step cross entropy loss of Section~\ref{sec:loss}. 
We perform data augmentation using random rotations of 10 degrees and horizontal flip with a 0.5 probability. All the models were implemented in Pytorch~\cite{PyTorch}, except for the RefineNet-based one, for which we used the MatConvNet implementation released by its authors. We will make our code publicly available upon acceptance of the paper, for both version of our Recurrent U-Net, along with the U-Net-based baselines.

\parag{Metrics.}
We report the same metrics as in~\cite{Urooj18}, that is, the mean intersection of union (mIoU), mean recall and mean precision. When validating the model parameters, however, we restrict ourselves to the mIoU, which is the standard metric used to evaluate semantic segmentation techniques and encompasses both precision and recall.

\subsection{Parameter Validation}
\label{sec:abl_para}
As shown in Fig.~\ref{fig:model_arch} and discussed in Section~\ref{sec:r-unet}, our recurrent unit can encompass different portions of the U-Net architecture, corresponding to different values of $\ell \in  \{0,\dots,4\}$ in our formalism. Furthermore, we need to initialize the segmentation mask $s_0$ and the hidden state $h_0$ for the first recursive iteration, and to define the parameter $\alpha$ in Eq.~\ref{eq:rel_weight}. Because of limited resources, we could not perform a complete grid search on all these hyper-parameters jointly. We therefore initialized $s_0$ and $h_0$ with the same value, which could be either all 0s or all 1s, and we separated the validation of these values and $\alpha$ from that of the choice of $\ell$. We used the EYTH validation set to choose these parameters. We then used the same values in all our experiments. Note that, while our approach to validation would clearly benefit from using more computational resources, we will show in Section~\ref{sec:SOTA} that it is sufficient to outperform the state of the art.

\begin{table}[!htb]
	\centering
	\resizebox{\linewidth}{!}{%
	\setlength{\tabcolsep}{2pt}
	\begin{tabular}[width=\textwidth]{r c  @{}cc c @{}cc  c @{}cc c @{}cc c @{}cc }
		\toprule
		& &\multicolumn{2}{c}{\TourS} & & \multicolumn{2}{c}{\TourD}  & & \multicolumn{2}{c}{\TrecS} & & \multicolumn{2}{c}{\TrecL } & & \multicolumn{2}{c}{\TrecM} \\
		\cmidrule(r){3-4}  	\cmidrule(r){6-7}   		\cmidrule(r){9-10}  \cmidrule(r){12-13}
		\cmidrule(r){15-16}
		Param. &\phantomtwo & $\alpha{:}0.4$  &  $\alpha{:}1.0$ & & $\alpha{:}0.4$  &  $\alpha{:}1.0$ & & $\alpha{:}0.4$  &  $\alpha{:}1.0$ & & $\alpha{:}0.4$  &  $\alpha{:}1.0$& & $\alpha{:}0.4$  &  $\alpha{:}1.0$\\
		\midrule 
		$s_0/h_0{:}0$ && 0.835 & 0.832  &&  0.816 & 0.833 &&  0.823 & 0.823 && 0.809 & 0.8274  && 0.826 & 0.825 \\
		$s_0/h_0{:}1$ && \textbf{0.837}  & 0.827 && \textbf{0.836}&0.832 && \textbf{0.831} & 0.825 &&0.816 & \textbf{0.829} && 0.823 & \textbf{0.831} \\ 
		\bottomrule
	\end{tabular}
	}
	\caption{\textbf{Validating $\mathbf{\alpha}$, $\mathbf{s_0}$ and $\mathbf{h_0}$.} We report the mean IoU on the EYTH validation set. In general, the methods benefit from initializing $s_0$ and/or $h_0$ to all 1s with $\alpha{=}0.4$.}
	\label{tab:params}
\end{table}

\parag{Choosing $\mathbf{\alpha}$, $\mathbf{s_0}$, and $\mathbf{h_0}$.} 
As mentioned above, we take $s_0$ and $h_0$ to have the same value of either all 0s or all 1s. Furthermore, we restrict $\alpha \in \{1, 0.4\}$, with the former giving the same weight to all recursive iterations, while the latter puts more emphasis on the later ones. We report the results in Table~\ref{tab:params}. 
For both versions of our approach, we used $\ell=0$ for the validation tests.
In general, we observe that all methods benefit from initializing $s_0/h_0$ to all 1s. In this case,  \ourS{}, \ourD{}, and \recS{} tend to perform best for $\alpha=0.4$. In the remaining experiments, we will use the hyper-parameter values that gave the best results for each method.

\begin{figure}[!t]
	\centering
	\includegraphics[width=\linewidth, height=4cm]{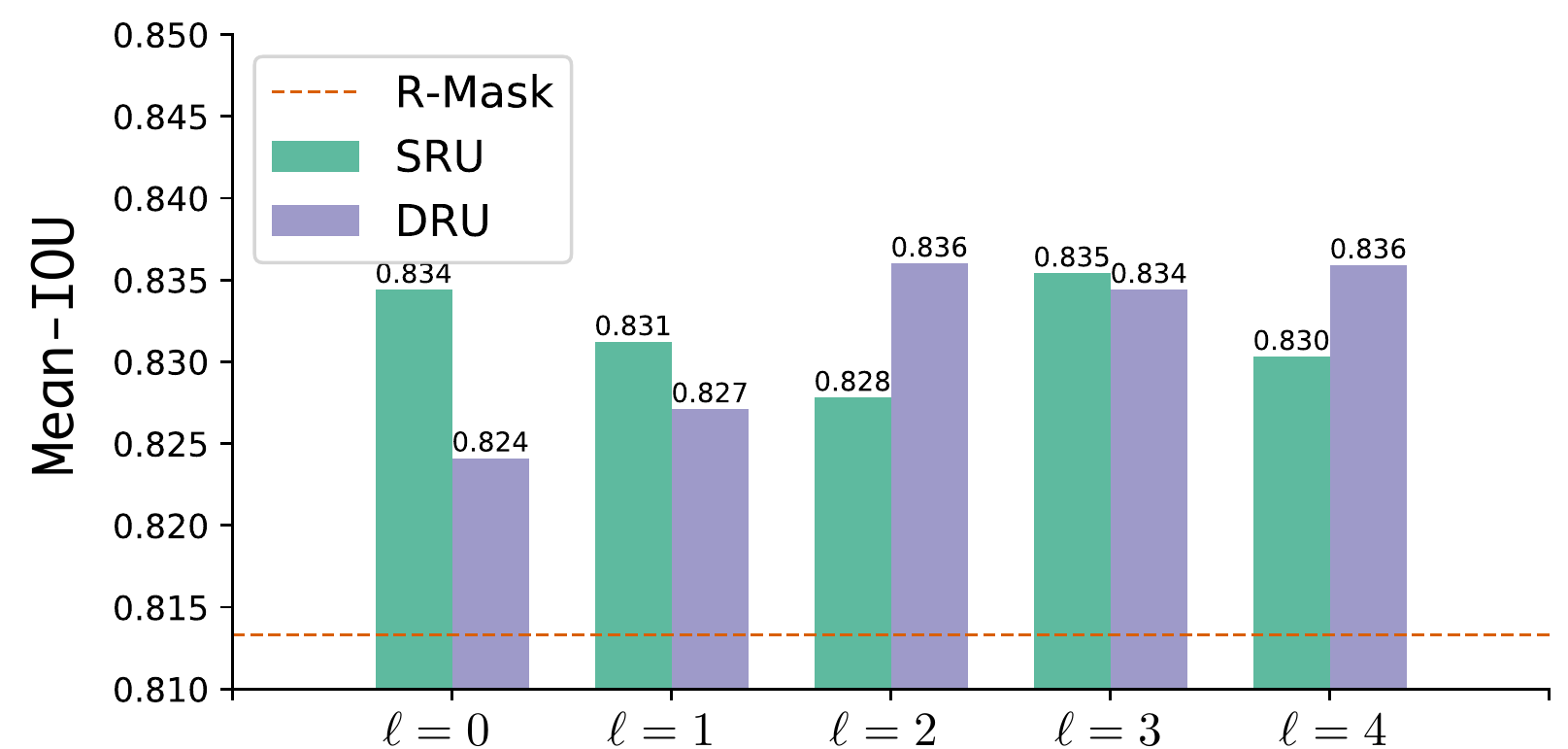}
	\caption{{\bf Validating $\mathbf{\ell.}$} We report the mean IoU on the EYTH validation set for different versions of our R-UNet, with either SRU or DRU. Note that all variants outperform the \recS baseline.}
	\label{fig:options}
\end{figure}

\begin{table*}[t]
	\centering
	\resizebox{\textwidth}{!}{%
		\setlength{\tabcolsep}{2pt}
		\begin{tabular}[width=\textwidth]{l| c @{}ccc c @{}ccc  c @{}ccc c @{}ccc c @{}ccc}
			\toprule
			Model &  & \multicolumn{3}{c}{EYTH~\cite{Urooj18}} & & \multicolumn{3}{c}{GTEA~\cite{Fathi11}}  & & \multicolumn{3}{c}{EgoHand~\cite{Bambach15}} && \multicolumn{3}{c}{HOF~\cite{Urooj18}} && \multicolumn{3}{c}{KBH} \\
			\cmidrule(r){1-1} 		\cmidrule(r){3-5}  		\cmidrule(r){7-9}   		\cmidrule(r){11-13}   		\cmidrule(r){15-17}   		\cmidrule(r){19-21}  
			\phantomone & & mIOU  & mRec  & mPrec & & mIOU  & mRec  & mPrec & & mIOU  & mRec  & mPrec & & mIOU  & mRec  & mPrec & & mIOU  & mRec  & mPrec \\
			\midrule
			\textit{One-shot baselines}\\
			\TrefN~\cite{Urooj18} & & 0.688 & 0.776 &  0.853 & &  0.821 & 0.869 & 0.928 & &  0.814 & 0.919 & 0.879 & &  \textbf{0.766} &\textbf{ 0.882} &  \textbf{0.859} &&  0.865 & 0.954 & 0.921 \\
			\TUnet-B~\cite{Ronneberger15} && 0.803 &0.912 & 0.830 &  & 0.950 & 0.973 & 0.975 & & 0.815 & 0.869 & 0.876 && 0.694 & 0.867 & 0.778 && 0.870 & 0.943 & 0.911 \\
			\TUnet-G &&0.837 & 0.928 & 0.883 & & 0.952 & 0.977 & 0.980 & &0.837 & 0.895 & 0.899 && 0.621 & 0.741 & 0.712  && 0.905 & 0.949 &0.948 \\
			\midrule
			\textit{Rec. baselines} \\
			\TrecM~\cite{Poudel16} && 0.827 &0.920&0.877 &   & 0.924 &0.979 & 0.976 && 0.828 &0.894 &0.905 && 0.654 &0.733 & 0.796& & 0.845 & 0.924 & 0.898 \\
			\TrecL~\cite{Valipour17} & & 0.838 &0.920  &0.894 &&0.957& 0.975 & 0.980 &&0.831&0.906 &0.897&& 0.674& 0.807 & 0.752  && 0.870 & 0.930	 & 0.924  \\
			\TrecS~\cite{Mosinska18} & & 0.827 & 0.918 & 0.864 & & 0.952 & 0.975 & 0.976 && 0.858 & 0.909 & 0.931 & & 0.693 & 0.833 & 0.704 & &  0.905 & 0.951 & 0.944  \\
			\midrule
			\multicolumn{2}{l} {\textit{Ours at layer ($\ell$) }} & \\
			\TourS (0)  & & 0.844 &0.924&0.890&& \textbf{0.960} & 0.976 & \textbf{0.981} & & 0.862 &0.913 & 0.932 & & 0.712 & 0.844 & 0.764 && 0.930 & 0.968 & 0.957\\
			\TourS (3)  && 0.845 & 0.931 &0.891 && 	0.956 & 0.977 & 0.982 &&0.864 &0.913 & 0.933 && 0.699 & 0.864 & 0.773  & & 0.921 &0.964& 0.951 \\
			\TourD (4) & & \textbf{0.849} & 	\textbf{0.926} &\textbf{ 0.900} &  & 0.958 &\textbf{0.978} & 0.977&  & \textbf{0.873 }& \textbf{0.924} & \textbf{0.935} & & 0.709 & 0.866 & 0.774 & &  \textbf{0.935} & \textbf{0.980} & \textbf{0.970}\\
			\bottomrule
		\end{tabular}
	}
	\caption{\small \textbf{Comparing against the state of the art. }  \ourD(4) does better overall, with \ourS(0) a close second. Generally speaking all recurrent methods do better than \refN{}, which represents the state of the art, on all datasets except HOF. We attribute this to HOF being too small for optimal performance without pre-training, as in  \refN{}. 
}
	\label{tab:hand}
\end{table*}

\parag{Choosing $\mathbf{\ell}$.}
We then evaluate the influence of $\ell$, which defines the portion of the \Unet{} used for internal state recurrence, as depicted by Fig.~\ref{fig:model_arch}(a). The respective accuracies of the \ourS~and \ourD~versions of our approach for different values of $\ell \in \{0,\dots,4\}$ are given in Fig.~\ref{fig:options}. From now on, we will denote by \ourS($\ell$) and \ourD($\ell$) the corresponding variant of our method that uses that value of $\ell$. 
For \ourS, $\ell=3$ and $\ell=0$ yield results that are within 0.1 mean IoU of each other. To favor incorporating as large a portion of the U-Net in out recurrent unit, so as to give a much flexibility to the network as possible to refine the segmentation,
we choose $\ell=0$ and report the results of \ourS{}(0). For \ourD{}, we obtain virtually the same results for $\ell \in \{2,3,4\}$. Since $\ell = 4$ requires incorporating fewer encoder-decoder layers in the recurrent unit, and thus duplicating fewer layers, we report the results of  \ourD{}(4) in the following experiments.

\parag{Choosing the Number of Recursive Iterations.}

\begin{figure}[htbp]
	\includegraphics[width=\linewidth]{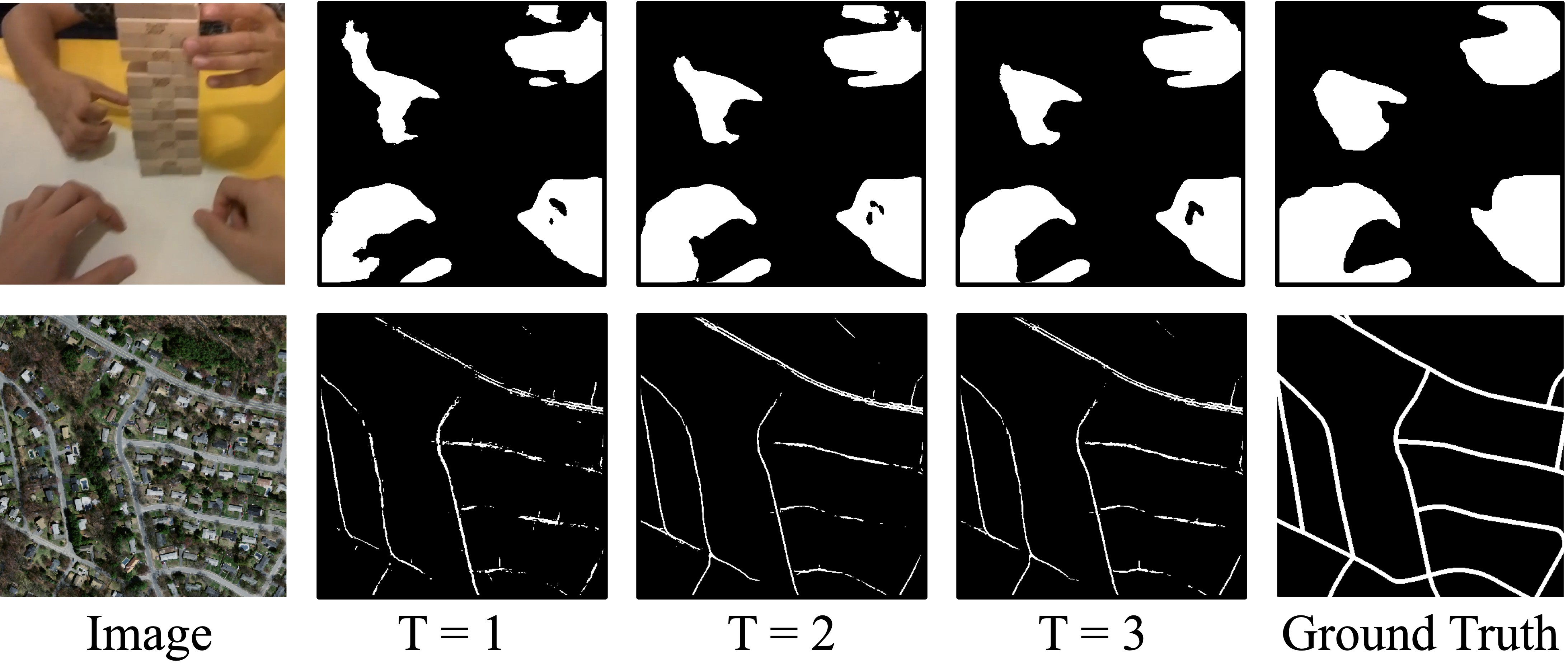}
	\caption{\small {\bf Recursive refinement.} Road and hand images; segmentation results after 1, 2, and 3 iterations; ground truth. Note the progressive refinement and the holes of the hands and roads being filled recursively.}
	\label{fig:recSteps}
\end{figure}

All the previous experiments were performed using $T=3$ recurrent iterations. Fig.~\ref{fig:recSteps} depicts the segmentation results on hands and roads after 1, 2, and 3 such iterations.  This number can also be varied and we plot in Table~\ref{tab:rec}  the accuracy as a function of $T$.  In practice, the accuracy typically increases until $T=3$ and then plateaus. 

\begin{table}
	\centering
	\resizebox{0.95\linewidth}{!} {
	\begin{tabular}{cc c c c c}
		\toprule
		Model 	&&\multicolumn{4}{c}{mIOU at Steps}\\
		\cmidrule{3-6}
		&   & $T= 1$ &  $T= 2$ &  $T= 3$ &  $T= 4$ \\
		\midrule
		\TourS{}(0)  && 0.816 & 0.8265 & 0.837 & 0.826 \\ 
		\TourD{}(4)  && 0.830 & 0.832 & 0.838 & 0.838 \\ 
		\bottomrule
	\end{tabular} 
	}
	\caption{\small \textbf{Influence of the number recurrent steps $T$.}}
	\label{tab:rec}
\end{table}

\subsection{Comparison to the State of the Art}
\label{sec:SOTA}

\begin{figure*}
	\includegraphics[width=\linewidth]{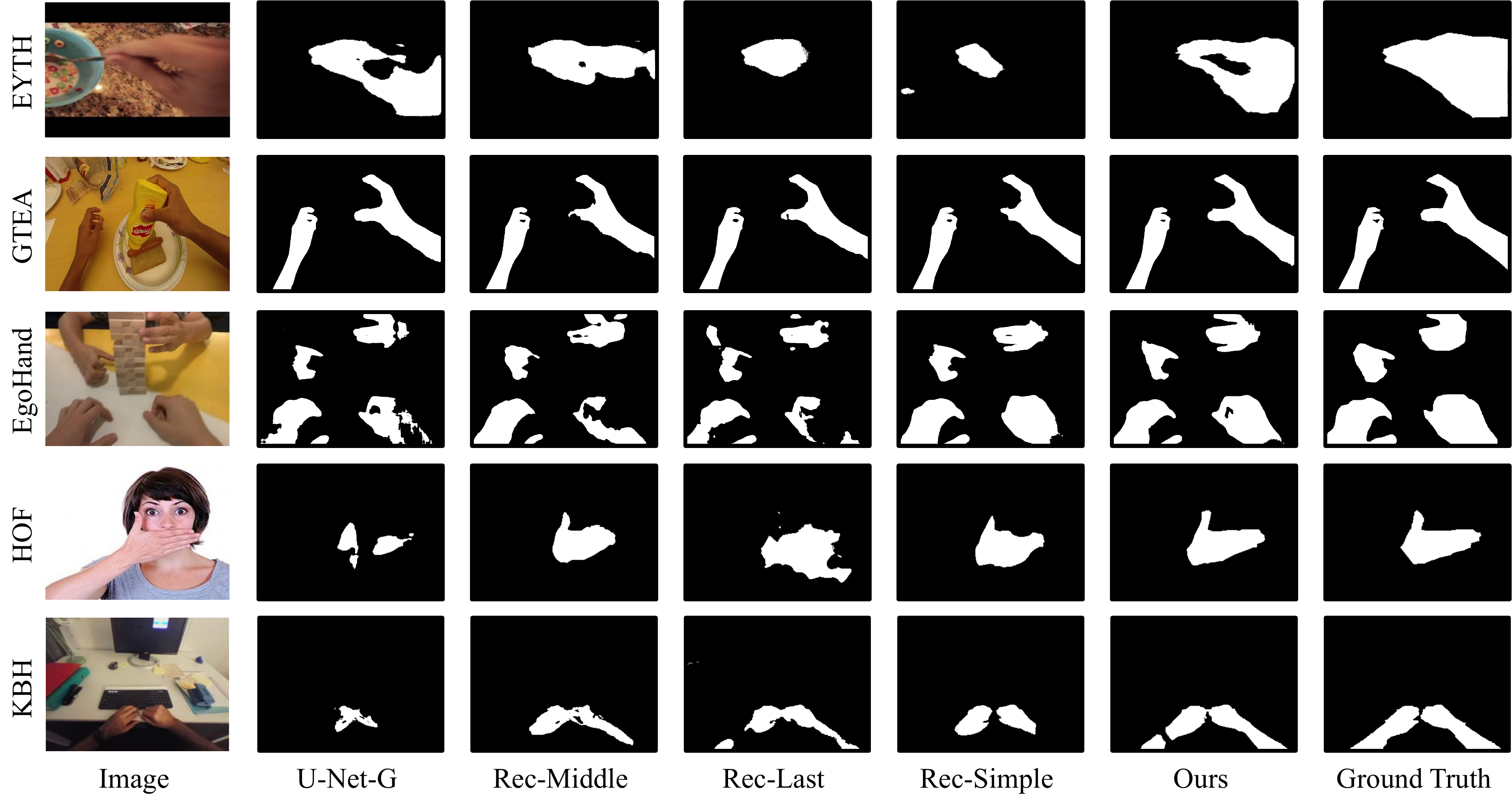}
	\caption{\small {\bf Example predictions on hand segmentation datasets.}}
	\label{fig:hand_vis}
\end{figure*}

We now compare the two versions of our approach to the baselines introduced above on the 5 benchmarks datasets we also described above. Table~\ref{tab:hand} summarizes our results. 

Overall, all the recurrent methods significantly outperform the two one-shot ones, \refN{} and \Unet{}, on four of the five datasets, which is remarkable given than \refN{} represents the current state of the art. Among the recurrent ones, \ourD(4) and \ourS(0) clearly dominate with \ourD(4) usually outperforming \ourS(0) by a small margin. Note that, even though \ourD(4)  as depicted by Fig.~\ref{fig:model_arch}(a)  looks superficially similar to \recM{}, they are quite different because \ourD{} takes the segmentation mask as input and relies our the new DRU gate, as discussed at the end of Section~\ref{sec:r-unet} and in Section~\ref{sec:dru}.  To confirm this, we evaluated a simplified version \ourD(4) in which we removed the segmentation mask from the input. The validation mIOU on EYTH decreased from 0.8355 to 0.8256 but remains better than that of \recM{} that is 0.814.

The one exception is the HOF dataset in which \refN{} still does best and we come second.   We believe that this can be attributed to HOF being the smallest dataset, with only 198 training images. Under such conditions,  \refN{}  strongly benefits from exploiting a ResNet-101 backbone that was pre-trained on PASCAL person parts~\cite{Chen14a}, instead of training for scratch as we do.

In Fig.~\ref{fig:hand_vis}, we provide a visual comparison of our results with those of the baselines. Note that our method yields accurate segmentation in diverse conditions, such as with hands close to the camera, multiple hands, hands over other skin regions, and low contrast images in our KBH dataset. By contrast, the baselines all fail in at least one of these scenarios. Interestingly,  our method sometimes yields a seemingly a more accurate segmentation than the ground-truth ones. For example, in our EYTH result at the top of Fig.~\ref{fig:hand_vis}, the gap between the thumb and index fingers is correctly found whereas it is missing from the ground truth. Likewise, for KBH at the bottom, the watch band is correctly identified as not being part of the arm even though it is labeled as such in the ground truth. 

\subsection{Generalizing to Another Segmentation Task}

This paper focuses on hand segmentation but our approach is generic and could be extended to other tasks. To demonstrate this, we employ the Massachusetts Roads dataset~\cite{Mnih13} that was used in~\cite{Mosinska18}. Here the task is to detect roads in high resolution satellite images, and we compare ourselves to the same baselines as before, using all the same hyperparameters as before. We report the same metrics as before, with two additional ones: precision-recall breaking point~(P/R) and F1-score. The cutting threshold for all metrics is set to 0.5 except for P/R. As baselines, we ignore \Unet-B since the corresponding \Unet-G is consistently better. With the same setting as hand-segmentation \recL~could not converge. As shown in Table~\ref{tab:road}, we also outperform all of the baselines on this task. 

Note that a P/R value of 0.773 is reported in~\cite{Mosinska18}, while our best model achieves 0.757. We consider it as a remarkable achievement keeping in mind that they used an-additional topology-aware loss and a much larger \Unet{}  based on 3 layers of a VGG19 pre-trained on ImageNet. In other words, we use 8 times fewer parameters than they do. This suggests that there is room for further improving our method by using a larger network and performing some pre-training with extra data.

\begin{table}[t]
	\centering
	\setlength{\tabcolsep}{5pt}
	\resizebox{\linewidth}{!}{
	\begin{tabular}[width=\linewidth] {l | c @{}ccccc}
	\toprule
	Models && mIOU & mRec & mPrec & P/R & F1 \\
	\midrule
	\TUnet-G~\cite{Ronneberger15} && 0.479 & 0.639 &0.502 & 0.642 & 0.563 \\
	\TrecM~\cite{Valipour17} && 0.494 & 0.767 & 0.518 & 0.660 &  0.574 \\
	\TrecS~\cite{Mosinska18} && 0.534 & 0.802 & 0.559 & 0.723 & 0.659 \\
	\midrule
	\TourS (0)  && 0.553 & 0.804 & 0.577& 0.727 & 0.671 \\
	\TourD (4)  && \textbf{0.560} &\textbf{ 0.865}& \textbf{0.583} & \textbf{0.757} & \textbf{0.691 } \\
	\bottomrule
\end{tabular}
}
	\caption{ \textbf{Road segmentation results.}}
	\label{tab:road}
	
\end{table}


\section{Conclusion}
We have introduced a recurrent approach to hand segmentation that combines a recursion on the segmentation mask and on the network's internal state. To this end, we have developed two different recurrent units that incorporate several encoding and decoding layers of the segmentation network. Our experimental results have demonstrated that our approach outperforms state-of-the-art one-shot segmentation methods and recurrent baselines that either only act on the segmentation mask or simply insert a standard recurrent unit in the segmentation network. We have also evidenced that our strategy generalizes beyond the hand-segmentation task it was designed for. In the future, we will therefore focus on building it on other backbone networks and evaluating it on other semantic segmentation tasks.


{\small
\begin{thebibliography}{10}\itemsep=-1pt

\bibitem{HTCvive}
{HTC Vive Virtual Reality Toolkit}.
\newblock \url{https://www.vive.com/}.

\bibitem{Argyros04}
A.~Argyros and M.~Lourakis.
\newblock {Real-Time Tracking of Multiple Skin-Colored Objects with a Possibly
  Moving Camera}.
\newblock In {\em European Conference on Computer Vision}, pages 368--379,
  2004.

\bibitem{Ballas16}
N.~Ballas, L.~Yao, C.~Pal, and A.~Courville.
\newblock {Delving Deeper into Convolutional Networks for Learning Video
  Representations}.
\newblock {\em International Conference on Learning Representations}, 2016.

\bibitem{Bambach15}
S.~Bambach, S.~Lee, D.~J. Crandall, and C.~Yu.
\newblock {Lending a Hand: Detecting Hands and Recognizing Activities in
  Complex Egocentric Interactions}.
\newblock In {\em International Conference on Computer Vision}, pages
  1949--1957, 2015.

\bibitem{Betancourt17}
A.~Betancourt, P.~Morerio, E.~Barakova, L.~Marcenaro, M.~Rauterberg, and
  C.~Regazzoni.
\newblock {Left/right Hand Segmentation in Egocentric Videos}.
\newblock 2017.

\bibitem{Chen14a}
X.~Chen, R.~Mottaghi, X.~Liu, S.~Fidler, R.~Urtasun, and A.~Yuille.
\newblock {Detect What You Can: Detecting and Representing Objects Using
  Holistic Models and Body Parts}.
\newblock In {\em Conference on Computer Vision and Pattern Recognition}, 2014.

\bibitem{Cho14c}
K.~Cho, B.~van Merrienboer, D.~Bahdanau, and Y.~Bengio.
\newblock {On the Properties of Neural Machine Translation: Encoder-Decoder
  Approaches}.
\newblock {\em arXiv Preprint}, 2014.

\bibitem{Fathi11}
A.~Fathi, A.~Farhadi, and J.~M. Rehg.
\newblock {Understanding Egocentric Activities}.
\newblock In {\em International Conference on Computer Vision}, pages 407--414,
  2011.

\bibitem{Goodfellow16}
I.~Goodfellow, Y.~Bengio, and A.~Courville.
\newblock {\em {Deep Learning}}.
\newblock MIT Press, 2016.

\bibitem{Hayman03}
E.~Hayman and J.-O. Eklundh.
\newblock {Statistical Background Subtraction for a Mobile Observer}.
\newblock In {\em International Conference on Computer Vision}, 2003.

\bibitem{He15}
K.~He, X.~Zhang, R.~Ren, and J.~Sun.
\newblock {Delving Deep into Rectifiers: Surpassing Human-Level Performance on
  Imagenet Classification}.
\newblock In {\em International Conference on Computer Vision}, 2015.

\bibitem{Jones02}
M.~J. Jones and J.~M. Rehg.
\newblock {Statistical Color Models with Application to Skin Detection}.
\newblock {\em International Journal of Computer Vision}, pages 81--96, 2002.

\bibitem{Kakumanu07}
P.~Kakumanu, S.~Makrogiannis, and N.~Bourbakis.
\newblock {A Survey of Skin-Color Modeling and Detection Methods}.
\newblock {\em Pattern Recognition}, 40(3):1106--1122, 2007.

\bibitem{Kolsch05}
M.~Kolsch and M.~Turk.
\newblock {Hand Tracking with Flocks of Features}.
\newblock In {\em Conference on Computer Vision and Pattern Recognition}, 2005.

\bibitem{Li13d}
C.~Li and K.~M. Kitani.
\newblock {Pixel-Level Hand Detection in Ego-Centric Videos}.
\newblock In {\em Conference on Computer Vision and Pattern Recognition}, 2013.

\bibitem{Lin17c}
G.~Lin, A.~Milan, C.~Shen, and I.~Reid.
\newblock {Refinenet: Multi-Path Refinement Networks for High-Resolution
  Semantic Segmentation}.
\newblock In {\em Conference on Computer Vision and Pattern Recognition}, 2017.

\bibitem{Long15a}
J.~Long, E.~Shelhamer, and T.~Darrell.
\newblock {Fully Convolutional Networks for Semantic Segmentation}.
\newblock In {\em Conference on Computer Vision and Pattern Recognition}, 2015.

\bibitem{Mnih13}
V.~Mnih.
\newblock {\em {Machine Learning for Aerial Image Labeling}}.
\newblock PhD thesis, University of Toronto, 2013.

\bibitem{Mosinska18}
A.~Mosinska, P.~Marquez-neila, M.~Kozinski, and P.~Fua.
\newblock {Beyond the Pixel-Wise Loss for Topology-Aware Delineation}.
\newblock In {\em Conference on Computer Vision and Pattern Recognition}, 2018.

\bibitem{Oikonomidis10}
I.~Oikonomidis, N.~Kyriazis, and A.~A. Argyros.
\newblock {Markerless and Efficient 26-DOF Hand Pose Recovery}.
\newblock In {\em Asian Conference on Computer Vision}, 2010.

\bibitem{PyTorch}
A.~Paszke, S.~Gross, S.~Chintala, G.~Chanan, E.~Yang, Z.~Devito, Z.~Lin,
  A.~Desmaison, L.~Antiga, and A.~Lerer.
\newblock {Automatic Differentiation in Pytorch}.
\newblock In {\em Advances in Neural Information Processing Systems}, 2017.

\bibitem{Pinheiro14}
P.~Pinheiro and R.~Collobert.
\newblock {Recurrent Neural Networks for Scene Labelling}.
\newblock In {\em International Conference on Machine Learning}, 2014.

\bibitem{Poudel16}
R.~P. Poudel, P.~Lamata, and G.~Montana.
\newblock {Recurrent Fully Convolutional Neural Networks for Multi-Slice MRI
  Cardiac Segmentation}.
\newblock In {\em Reconstruction, Segmentation, and Analysis of Medical
  Images}, pages 83--94. Springer, 2016.

\bibitem{Purves01}
D.~Purves, G.~J. Augustine, D.~Fitzpatrick, L.~C. Katz, A.-S. LaMantia, J.~O.
  McNamara, S.~M. Williams, et~al.
\newblock Types of eye movements and their functions.
\newblock {\em Neuroscience}, pages 361--390, 2001.

\bibitem{Romera16}
B.~Romera-Paredes and P.~H.~S. Torr.
\newblock {Recurrent Instance Segmentation}.
\newblock In {\em European Conference on Computer Vision}, pages 312--329,
  2016.

\bibitem{Ronneberger15}
O.~Ronneberger, P.~Fischer, and T.~Brox.
\newblock {{U-Net}: Convolutional Networks for Biomedical Image Segmentation}.
\newblock In {\em Conference on Medical Image Computing and Computer Assisted
  Intervention}, 2015.

\bibitem{Rother04}
C.~Rother, V.~Kolmogorov, and A.~Blake.
\newblock {"{GrabCut}" - Interactive Foreground Extraction Using Iterated Graph
  Cuts}.
\newblock In {\em ACM SIGGRAPH}, pages 309--314, 2004.

\bibitem{Sheikh09}
Y.~Sheikh, O.~Javed, and T.~Kanade.
\newblock {Background Subtraction for Freely Moving Cameras}.
\newblock In {\em International Conference on Computer Vision}, 2009.

\bibitem{Simonyan15}
K.~Simonyan and A.~Zisserman.
\newblock {Very Deep Convolutional Networks for Large-Scale Image Recognition}.
\newblock In {\em International Conference on Learning Representations}, 2015.

\bibitem{Stenger01a}
B.~Stenger, P.~Mendonca, and R.~Cipolla.
\newblock {Model-Based 3D Tracking of an Articulated Hand}.
\newblock In {\em Conference on Computer Vision and Pattern Recognition}, pages
  310--315, December 2001.

\bibitem{Sudderth04}
E.~B. Sudderth, M.~I. Mandel, W.~T. Freeman, and A.~S. Willsky.
\newblock {Visual Hand Tracking Using Nonparametric Belief Propagation}.
\newblock In {\em Conference on Computer Vision and Pattern Recognition}, 2004.

\bibitem{Urooj18}
A.~Urooj and A.~Borji.
\newblock {Analysis of Hand Segmentation in the Wild}.
\newblock In {\em Conference on Computer Vision and Pattern Recognition}, 2018.

\bibitem{Valipour17}
S.~Valipour, M.~Siam, M.~Jagersand, and N.~Ray.
\newblock {Recurrent Fully Convolutional Networks for Video Segmentation}.
\newblock In {\em IEEE Winter Conference on Applications of Computer Vision},
  2017.

\bibitem{Wu18a}
Y.~Wu and K.~He.
\newblock {Group Normalization}.
\newblock In {\em European Conference on Computer Vision}, 2018.

\end{thebibliography}

}
\end{document}